\documentclass[11pt]{article}
\usepackage[final]{acl}             
\usepackage{times}
\usepackage{latexsym}
\usepackage[T1]{fontenc}
\usepackage[utf8]{inputenc}
\usepackage{microtype}
\usepackage{graphicx}
\usepackage{booktabs}
\usepackage{amsmath}
\usepackage{multirow}
\usepackage{tikz}
\usetikzlibrary{shapes,arrows,positioning,fit}
\usepackage{orcidlink}

\title{UTS at PsyDefDetect: Multi-Agent Councils and\\Absence-Based Reasoning for Defense Mechanism Classification}

\author{
  Dima Galat\,\orcidlink{0000-0003-3825-2142}
  \and
  Marian-Andrei Rizoiu\,\orcidlink{0000-0003-0381-669X} \\
  University of Technology Sydney
}

\begin{document}
\maketitle

\begin{abstract}
This paper describes our system for classifying psychological defense mechanisms in emotional support dialogues using the Defense Mechanism Rating Scales (DMRS), placing second (F1 0.406) among 64 teams.\footnote{Code available at \url{https://github.com/dimagalat/bionlp2026}}
A central insight is that defense mechanisms are defined by what is \emph{absent}: missing affect, blocked cognition, denied reality.
We encode this as an \emph{affect-cognition integration spectrum} in prompt-level clinical rules, which account for the largest single gain (+11.4pp F1).

Our architecture is a multi-phase \emph{deliberative} council of Gemini 2.5 agents where class-specific advocates rate evidence strength rather than voting, achieving F1 0.382 with no fine-tuning---a top-5 result on its own.
We find, however, that the council is \emph{confidently wrong} about minority classes: 59--80\% of stable minority predictions are incorrect, driven by a systematic ``L7 attractor'' in which emotional content defaults to the majority class.
A targeted override ensemble from three fine-tuned Qwen3.5 models applies 16 overrides (+2.4pp), selected by a structured multi-agent system (builder, critic, regression guard) that produced a larger F1 gain in one iteration than 8 prior attempts combined.
\end{abstract}

\section{Introduction}
\label{sec:intro}

The BioNLP 2026 shared task \cite{na-etal-2026-psydefdetect,na-etal-2026-psydefconv} requires classifying target utterances in emotional support dialogues into 9 levels of the Defense Mechanism Rating Scales (DMRS; \citealt{perry1990dmrs}), a hierarchical clinical instrument ranging from action-based defenses (Level 1) to highly adaptive coping (Level 7).
The difficulty is that mechanisms are defined by psychological \emph{function}, not linguistic form: the same surface expression can indicate denial (Level 3), intellectualization (Level 6), or adaptive coping (Level 7) depending on context (illustrated in \S\ref{sec:task}).

The task exhibits severe class imbalance (Table~\ref{tab:distribution}): Level 7 (Highly Adaptive) comprises 51.9\% of training data, while Level 5 and Level 8 account for only 2.6\% and 1.5\%.
We call this the \emph{L7 attractor effect}: LLMs over-predict the majority class because emotional engagement in therapeutic dialogue looks like ``adaptive coping.''

Our system uses a multi-phase deliberative council (\S\ref{sec:council}), built on the Gemini 2.5 API \cite{gemini2023}: three specialist agents classify in parallel, class-specific advocates rate evidence strength, and a resolution stage adjudicates.
We frame the DMRS hierarchy as an affect-cognition integration spectrum: many defense mechanisms are defined by what is absent (missing affect, blocked cognition, denied reality), which requires reasoning about what should be present but is not.
A targeted override ensemble from three fine-tuned models applies 16 overrides to the council's predictions, achieving macro-F1 0.406 (2nd out of 21 registered teams, or 64 CodaBench entries).
We also flag a retrieval-leakage risk: same-dialogue exemplars in few-shot prompts inflate validation accuracy from 65\% to 97.7\% (\S\ref{sec:finetuned}).

\section{Task and Data}
\label{sec:task}

The dataset contains 1,864 training samples from 200 dialogues and 472 test samples from 189 dialogues.
All 189 test dialogue IDs overlap with training dialogues (different utterances from the same conversations), creating a retrieval leakage risk addressed in \S\ref{sec:finetuned}.
The official metric is macro-averaged F1; we write ``F1'' throughout to mean macro-F1.

\begin{table}[t]
\centering
\small
\begin{tabular}{clrr}
\toprule
\textbf{Level} & \textbf{Name} & \textbf{Train} & \textbf{\%} \\
\midrule
0 & No Defense / Neutral & 296 & 15.9 \\
1 & Action Defense & 108 & 5.8 \\
2 & Major Image-Distorting & 61 & 3.3 \\
3 & Disavowal & 99 & 5.3 \\
4 & Minor Image-Distorting & 84 & 4.5 \\
5 & Neurotic Defense & 48 & 2.6 \\
6 & Obsessional Defense & 172 & 9.2 \\
7 & Highly Adaptive & 968 & 51.9 \\
8 & Needs More Information & 28 & 1.5 \\
\bottomrule
\end{tabular}
\caption{DMRS class distribution in training data. Level 7 accounts for over half of all samples; the five lower-level defense classes (L1--L5) together comprise only 21.5\%.}
\label{tab:distribution}
\end{table}

The core challenge is illustrated by this training example: a speaker responds to ``How are you today?'' with \emph{``I'm OK. Just dealing with a lot of unknowns.''}
This reads like Level 7 (\emph{Suppression}, consciously managing distress). The ground truth is Level 6 (\emph{Isolation of Affect}): the speaker acknowledges difficulty cognitively (``a lot of unknowns'') but the expected emotional response is absent (``I'm OK''). The distinction turns on what is absent from the utterance, not what is present.

\section{System Architecture}
\label{sec:architecture}

\subsection{Multi-Phase Deliberative Council}
\label{sec:council}

Most LLM ensembles aggregate votes. A \emph{deliberative} council instead evaluates \emph{evidence strength per candidate} through structured advocacy.
In our architecture, Phase 1 agents propose candidates with alternatives (not just top-1 predictions).
Phase 2 spawns class-specific advocates that rate fit as \textsc{strong}, \textsc{moderate}, or \textsc{weak}; each argues \emph{for} its assigned class using retrieved exemplars.
Phase 3 resolves via evidence quality (unique \textsc{strong} wins immediately; ties require pairwise comparison), not vote count.
Majority voting loses minority classes because L7 always outnumbers them; evidence-based resolution can select a minority class that receives \textsc{strong} even when the majority favors L7.

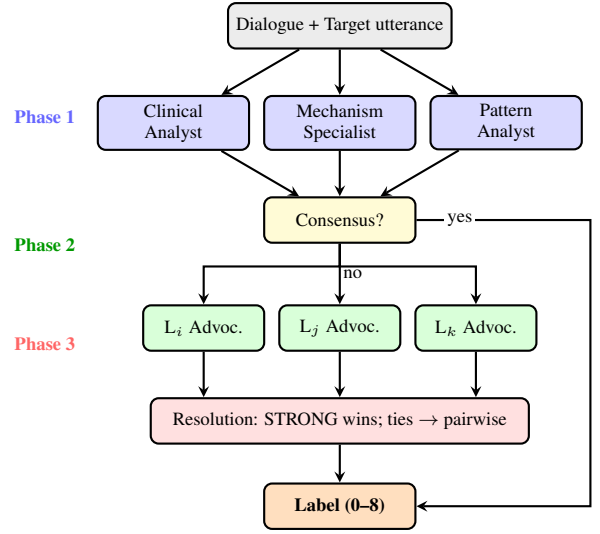
\begin{figure}[t]
\centering
\small
\begin{tikzpicture}[
    node distance=0.5cm,
    box/.style={draw, rounded corners=3pt, minimum width=2cm, minimum height=0.6cm, font=\scriptsize, align=center, thick},
    arr/.style={->, >=stealth, thick},
    lbl/.style={font=\scriptsize, fill=white, inner sep=1pt},
]
\node[box, fill=gray!15] (input) {Dialogue + Target utterance};

\node[box, fill=blue!15, below=0.6cm of input, xshift=-2.2cm] (ca) {Clinical\\Analyst};
\node[box, fill=blue!15, below=0.6cm of input] (ms) {Mechanism\\Specialist};
\node[box, fill=blue!15, below=0.6cm of input, xshift=2.2cm] (pa) {Pattern\\Analyst};

\node[font=\scriptsize\bfseries, color=blue!60] at (-3.9,-1.2) {Phase 1};

\node[box, fill=yellow!20, below=0.6cm of ms] (cons) {Consensus?};

\node[box, fill=green!15, below=0.8cm of cons, xshift=-1.8cm, minimum width=1.6cm] (adv1) {L$_i$ Advoc.};
\node[box, fill=green!15, below=0.8cm of cons, minimum width=1.6cm] (adv2) {L$_j$ Advoc.};
\node[box, fill=green!15, below=0.8cm of cons, xshift=1.8cm, minimum width=1.6cm] (adv3) {L$_k$ Advoc.};

\node[font=\scriptsize\bfseries, color=green!60!black] at (-3.9,-2.9) {Phase 2};

\node[box, fill=red!12, below=0.6cm of adv2, minimum width=5cm] (res) {Resolution: STRONG wins; ties $\to$ pairwise};

\node[font=\scriptsize\bfseries, color=red!60] at (-3.9,-4.2) {Phase 3};

\node[box, fill=orange!25, below=0.5cm of res] (out) {\textbf{Label (0--8)}};

\draw[arr] (input) -- (ca);
\draw[arr] (input) -- (ms);
\draw[arr] (input) -- (pa);
\draw[arr] (ca) -- (cons);
\draw[arr] (ms) -- (cons);
\draw[arr] (pa) -- (cons);

\draw[arr] (cons.east) -- ++(2.3,0) node[lbl, pos=0.25] {yes} |- (out.east);

\draw[arr] (cons) -- node[lbl, right] {no} (adv2);
\draw[arr] (cons.south) -- ++(0,-0.3) -| (adv1.north);
\draw[arr] (cons.south) -- ++(0,-0.3) -| (adv3.north);

\draw[arr] (adv1.south) -- (adv1 |- res.north);
\draw[arr] (adv2.south) -- (res.north);
\draw[arr] (adv3.south) -- (adv3 |- res.north);

\draw[arr] (res) -- (out);
\end{tikzpicture}
\caption{Council pipeline (3--10 LLM calls per sample). Phase 1: three specialist agents classify in parallel. If unanimous with high confidence, the pipeline exits early (3 calls). Otherwise, Phase 2 spawns class-specific advocates rating fit as \textsc{strong}/\textsc{moderate}/\textsc{weak} (2--5 calls). Phase 3 resolves via priority hierarchy (0--1 call).}
\label{fig:council}
\end{figure}

Figure~\ref{fig:council} illustrates the pipeline, which uses the Gemini 2.5 API (primarily Flash, with Pro for resolution) and requires 3--10 LLM calls per sample depending on consensus.

Formally, given a dialogue $d$ with target utterance $u$ and label space $\mathcal{Y} = \{0, \ldots, 8\}$, the council proceeds in three stages.
Three agents $a_1, a_2, a_3$ each produce a candidate set $C_i = \{(y_i, y'_i, p_i)\}$ (primary label, alternative, confidence).
Let $\mathcal{C} = \bigcup_i \{y_i, y'_i\}$ be the candidate pool.
If all primaries agree ($y_1 = y_2 = y_3 = y^*$) with $\sum_i \mathbf{1}[p_i \geq \tau] \geq 2$, the council returns $y^*$ immediately.
Otherwise, for each unique candidate $c \in \mathcal{C}$, a class-specific advocate $A_c$ produces a strength rating $s(c) \in \{\textsc{strong}, \textsc{moderate}, \textsc{weak}\}$.
A resolution function $R$ selects the final label via priority ordering: unique \textsc{strong} wins; ties are resolved by pairwise head-to-head comparison.
Phase details are in Appendix~\ref{app:council_details}.

\paragraph{Clinical Knowledge Encoding.}
We encode the DMRS hierarchy \cite{perry1990dmrs} as an \emph{affect-cognition integration spectrum}.
The discriminative question at each level is: what is the relationship between what the speaker knows and what they feel?
Cognition present but affect drained $\to$ L6 (Obsessional);
affect present but cognition blocked $\to$ L5 (Neurotic);
cognition distorts reality to manage affect $\to$ L2--4;
affect and cognition integrated $\to$ L7 (Adaptive).
The most impactful single test is \textbf{Reporting vs.\ Processing} (L6/L7): describing painful facts without proportional emotion is Isolation of Affect (L6), not adaptive coping (L7).
We complement this with five prompt-level rules: (1) 60+ DMRS-Q behavioral indicators per mechanism \cite{perry1990dmrs}; (2) \emph{Emotion $\neq$ Defense}: ``I feel sad'' is not a defense without distortion or transformation; (3) prefer lower-level (less mature) defenses when ambiguous; (4) a watchlist of 8 high-confusion class pairs with discriminative tests; and (5) an L7 verification gate requiring a named adaptive mechanism before permitting L7.
In ablation, the Gemini 2.5 Pro council without these clinical rules achieves F1 0.268 (Table~\ref{tab:results}); adding them raises F1 to 0.382 (+11.4pp).

\paragraph{Phase-Level Bottleneck.}
Even with clinical rules, the council's residual errors concentrate at \emph{resolution} rather than detection: the correct label enters Phase 2 as a candidate in 96\% of errors and receives a \textsc{moderate}+ rating in 76\%, but in 94\% of errors a wrong label (typically L7) also receives \textsc{strong}.
The minority-class signal exists; what is missing is a way to prevent L7 from winning the head-to-head, which motivates the override ensemble (\S\ref{sec:ensemble}).

\subsection{Retrieval and Fine-Tuned Models}
\label{sec:finetuned}

Few-shot examples are retrieved via TF-IDF with MMR \cite{carbonell1998mmr} for Phase 1 diversity and semantic re-ranking for Phase 2 within-class exemplars \cite{lewis2020rag}.
Dialogue-ID exclusion prevents same-conversation leakage; without it, council validation accuracy inflates from 65\% to 97.7\%.

We train three models via LoRA \cite{hu2022lora} in 4-bit quantization \cite{dettmers2023qlora} using Unsloth \cite{unsloth2024} and TRL \cite{trl2024}: Qwen3.5-9B (65.1\% val acc, strongest on L6/L1), Qwen3.5-35B-A3B MoE (61.7\%, strongest on L2/L3), and Qwen3.5-9B f1\_boost (62.5\%, strongest on L1/L2) \cite{qwen2025}.
All use \textbf{completion-only loss} (\texttt{train\_on\_responses\_only}): training on the full sequence wastes $>$99\% of gradient updates on dialogue auto-completion, and this single change improves accuracy from 25--55\% to 59--65\%.
Self-consistency inference \cite{wang2023selfconsistency} (multiple runs at temperature=0.3) provides per-sample confidence scores.
A separate pairwise L6/L7 resolver (Qwen2.5-7B, 97.8\% val accuracy) handles the dominant confusion pair.
Per-model details are in Appendix~\ref{app:model_details}.

\subsection{Ensemble Strategy}
\label{sec:ensemble}

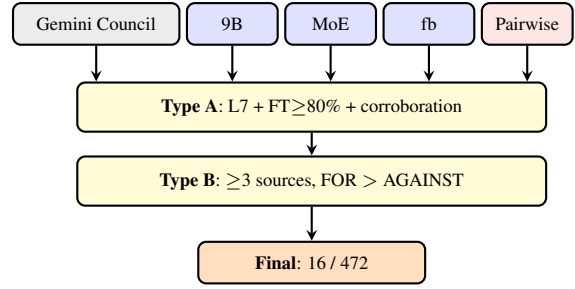
\begin{figure}[t]
\centering
\small
\begin{tikzpicture}[
    node distance=0.4cm,
    box/.style={draw, rounded corners=3pt, minimum height=0.55cm, font=\scriptsize, align=center, thick},
    src/.style={box, fill=blue!12, minimum width=1.3cm},
    gate/.style={box, fill=yellow!20, minimum width=6.5cm, minimum height=0.65cm},
    outbox/.style={box, fill=orange!25, minimum width=3cm},
    arr/.style={->, >=stealth, thick},
]
\node[src, minimum width=2.2cm, fill=gray!15] (council) at (-2.0, 0) {\scriptsize Gemini Council};

\node[src, minimum width=1.2cm] (ft1) at (-0.2, 0) {\scriptsize 9B};
\node[src, minimum width=1.2cm] (ft2) at (1.1, 0) {\scriptsize MoE};
\node[src, minimum width=1.2cm] (ft3) at (2.4, 0) {\scriptsize fb};
\node[src, minimum width=1.2cm, fill=red!10] (pw) at (3.7, 0) {\scriptsize Pairwise};

\node[gate, minimum width=6.2cm] at (0.85, -1.1) (gateA) {\textbf{Type A}: L7 + FT$\geq$80\% + corroboration};
\node[gate, minimum width=6.2cm, below=0.3cm of gateA] (gateB) {\textbf{Type B}: $\geq$3 sources, FOR $>$ AGAINST};

\node[outbox, below=0.45cm of gateB] (final) {\textbf{Final}: 16 / 472};

\draw[arr] (council.south) -- (council |- gateA.north);
\draw[arr] (ft1.south) -- (ft1 |- gateA.north);
\draw[arr] (ft2.south) -- (ft2 |- gateA.north);
\draw[arr] (ft3.south) -- (ft3 |- gateA.north);
\draw[arr] (pw.south) -- (pw |- gateA.north);

\draw[arr] (gateA.south) -- (gateB.north);
\draw[arr] (gateB.south) -- (final.north);
\end{tikzpicture}
\caption{Override ensemble. The council's predictions are checked against three fine-tuned models (20 self-consistency runs each) and a pairwise L6/L7 resolver. Type A corrects L7 over-predictions (7 overrides); Type B corrects minority confusions (9 overrides).}
\label{fig:ensemble}
\end{figure}

The ensemble applies minimal, high-confidence corrections to the council's predictions (Figure~\ref{fig:ensemble}).

\textbf{Type A} (L7$\to$minority, high risk): we override when a fine-tuned model achieves $\geq$80\% self-consistency for a minority class and a council rerun or pairwise resolver corroborates. If wrong, we lose a true L7 and add a false minority, a double penalty.
\textbf{Type B} (minority$\to$minority, lower risk): requires $\geq$3 of 6 independent sources to agree, with FOR~$>$~AGAINST. A credibility gate discards models with $<$15\% val recall on the target class.
Our final submission applies 7 Type A and 9 Type B corrections (3.4\% of predictions).

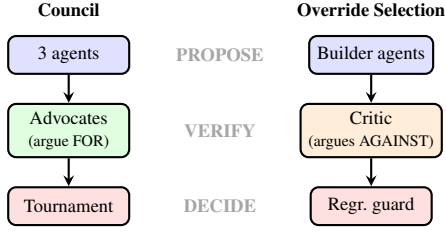
\begin{figure}[t]
\centering
\small
\begin{tikzpicture}[
    node distance=0.35cm,
    box/.style={draw, rounded corners=3pt, minimum height=0.5cm, minimum width=1.6cm, font=\scriptsize, align=center, thick},
    stage/.style={font=\scriptsize\bfseries, color=gray!70},
    arr/.style={->, >=stealth, thick},
]
\node[font=\scriptsize\bfseries] (t1) at (-2, 0.6) {Council};
\node[box, fill=blue!12] (p1) at (-2, 0) {3 agents};
\node[box, fill=green!12] (v1) at (-2, -1.0) {Advocates\\{\tiny (argue FOR)}};
\node[box, fill=red!12] (d1) at (-2, -2.0) {Tournament};

\node[font=\scriptsize\bfseries] (t2) at (2, 0.6) {Override Selection};
\node[box, fill=blue!12] (p2) at (2, 0) {Builder agents};
\node[box, fill=orange!15] (v2) at (2, -1.0) {Critic\\{\tiny (argues AGAINST)}};
\node[box, fill=red!12] (d2) at (2, -2.0) {Regr.\ guard};

\node[stage] at (0, 0) {PROPOSE};
\node[stage] at (0, -1.0) {VERIFY};
\node[stage] at (0, -2.0) {DECIDE};

\draw[arr] (p1) -- (v1);
\draw[arr] (v1) -- (d1);
\draw[arr] (p2) -- (v2);
\draw[arr] (v2) -- (d2);
\end{tikzpicture}
\caption{Both multi-agent systems follow a propose--verify--decide pattern. The council's verifiers are advocates (each argues for one class); the override system's verifier is an adversary (argues against all candidates).}
\label{fig:pattern}
\end{figure}

\paragraph{Agentic Override Selection.}
The override search space is combinatorial: 472 $\times$ 8 $=$ 3,776 candidates.
For the final submission, we decomposed override selection into three formal roles:
(1) parallel \emph{builder agents} scanning every sample against all sources with credibility gates;
(2) an independent \emph{critic agent} verifying every claim with separate data access; and
(3) a programmatic \emph{regression guard} hard-rejecting submissions below the evidence threshold.
This structured approach found 5 corrections that ad-hoc exploration had missed, pushing F1 from 0.393 to 0.406, a larger gain than the preceding 8 iterations combined.
The lesson: the value is in formal role decomposition, not automation; agents all the way down, but with structure at every level.
\paragraph{Propose--Verify--Decide.}
Both our council (\S\ref{sec:council}) and the override system follow a three-stage \emph{propose--verify--decide} pattern (Figure~\ref{fig:pattern}), differing in the verifier's stance: council advocates argue FOR each candidate class while the override critic argues AGAINST every proposal, and only candidates surviving its scrutiny pass the programmatic regression guard.
This mirrors the distinction between the generator--verifier paradigm \cite{cobbe2021verifiers} and adversarial debate \cite{irving2018debate}.
Separating proposer and verifier into distinct agents with independent data access prevents the confirmation bias of single-agent self-refine loops \cite{madaan2023selfrefine}.

\section{Key Findings}
\label{sec:findings}

\paragraph{L7 Attractor Effect.}
The council's L7 advocate rates \textsc{strong} 73\% of the time versus 32--43\% for other classes, causing 54\% of stable errors to be incorrect L7 predictions (Table~\ref{tab:confusions}).
The effect intensifies with dialogue length (15.0 turns for errors vs.\ 12.6 for correct) and minority class accuracy decays sharply with position: L1 drops from 100\% (early turns) to 16\% (late turns).

\begin{table}[t]
\centering
\small
\begin{tabular}{lrr}
\toprule
\textbf{True $\to$ Predicted} & \textbf{Count} & \textbf{\% Errors} \\
\midrule
L6 $\to$ L7 & 66 & 16.8 \\
L0 $\to$ L7 & 37 & 9.4 \\
L4 $\to$ L7 & 34 & 8.7 \\
L3 $\to$ L7 & 32 & 8.2 \\
L7 $\to$ L0 & 27 & 6.9 \\
L1 $\to$ L3 & 18 & 4.6 \\
\midrule
\emph{Any $\to$ L7} & \emph{213} & \emph{54.3} \\
\bottomrule
\end{tabular}
\caption{Top 6 stable error confusions across 3 council runs (392 total errors). Over half are incorrect L7 predictions; L6$\to$L7 alone accounts for 17\% of all errors.}
\label{tab:confusions}
\end{table}

\paragraph{Confidently Wrong.}
Three identical council runs show 22.2\% prediction instability, driven by asynchronous execution order in the multi-agent architecture even at temperature=0.
For minority classes L3, L4, L6, and L8, unstable predictions are paradoxically more accurate than stable ones (Table~\ref{tab:stability} in App.~\ref{app:findings}); between 59\% and 80\% of stable minority predictions are wrong.

\paragraph{Prompt Overfitting.}
Every prompt modification that improved training-set F1 degraded test performance (Table~\ref{tab:overfitting}).
We hypothesize this reflects prompt overfitting: iteratively tuning prompts against training-set metrics acts as implicit gradient descent with no generalization check.

\begin{table}[t]
\centering
\small
\begin{tabular}{lrr}
\toprule
\textbf{Modification} & \textbf{$\Delta$ Train} & \textbf{$\Delta$ Test} \\
\midrule
L7 advocate rigor & +1.2pp & $-$3.6pp \\
Dialogue position metadata & +0.8pp & $-$2.1pp \\
Training exemplars in prompts & +1.5pp & $-$3.6pp \\
Few-shot $k{=}3 \to 5$ & +1.5pp & $-$1.4pp \\
Length bias warning & +0.5pp & $-$2.0pp \\
\bottomrule
\end{tabular}
\caption{The overfitting paradox: modifications improving training F1 consistently degrade test F1.}
\label{tab:overfitting}
\end{table}

\paragraph{Additional findings.}
TF-IDF with MMR retrieves 38\% L7 examples versus 42--46\% for all semantic variants, because its lack of semantic understanding prevents emotional-content clustering, a useful property for diversity-dependent classification.
Completion-only loss is the most impactful fine-tuning intervention: it improves accuracy from 25--55\% to 59--65\%.

\paragraph{Override Selection Lessons.}
Beyond architecture, our 9 submissions yielded replicable patterns for ensemble correction on imbalanced classification, four of which are not derivable from individual model accuracy.
\textbf{Prediction volume signals per-sample reliability.} Under-predictors (9B predicts L2 for 11 samples vs.\ ${\sim}16$ expected) carry higher precision; over-predictors (f1\_boost: 31 vs.\ ${\sim}16$) inflate false positives, so we weight votes by this volume discount.
\textbf{Architectural independence is necessary but not sufficient.} A Gemma4-26B-A4B \cite{gemma2024} agreeing with the Qwen3.5 ensemble only 69--74\% of the time (the most independent source we trained) never improved test F1, because per-class accuracy was too low for its disagreements to carry signal.
\textbf{Override count vs.\ F1 is sharply non-monotonic.} 75 overrides $\to$ F1 .367; 16 $\to$ .406; 21 $\to$ .399; the peak lives at a narrow intersection of evidence quality and quantity.
\textbf{Regression guards must be programmatic.} Advisory warnings (``this override has weak evidence'') were ignored by human operators and LLM agents alike; only hard-rejecting submissions failing structural checks prevented regressions.

\section{Results}
\label{sec:results}

\begin{table}[t]
\centering
\small
\begin{tabular}{lcccc}
\toprule
\textbf{System} & \textbf{Acc} & \textbf{P} & \textbf{R} & \textbf{F1} \\
\midrule
Council (no clin.\ rules) & .653 & .395 & .249 & .268 \\
Council baseline & .663 & .473 & .364 & .382 \\
\quad + 7 Type A overrides & .670 & .477 & .377 & .391 \\
\quad + 9 Type B overrides & \textbf{.674} & \textbf{.461} & \textbf{.388} & \textbf{.406} \\
\bottomrule
\end{tabular}
\caption{Test set results. Clinical knowledge rules account for +11.4pp F1 over the unconfigured council; the override ensemble adds +2.4pp through 16 overrides.}
\label{tab:results}
\end{table}

\begin{table}[t]
\centering
\small
\begin{tabular}{lcccc}
\toprule
\textbf{Team} & \textbf{Acc} & \textbf{P} & \textbf{R} & \textbf{F1} \\
\midrule
N\"urnberg NLP & \textbf{.701} & .451 & \textbf{.404} & \textbf{.420} \\
UTS (ours) & .674 & \textbf{.461} & .388 & .406 \\
PerceptionLab & .674 & .426 & .409 & .396 \\
LinguIUTics & .642 & .400 & .396 & .392 \\
LDI Lab & .636 & .377 & .389 & .371 \\
\bottomrule
\end{tabular}
\caption{Top 5 on the shared task leaderboard (21 registered teams). Our system has the highest precision.}
\label{tab:leaderboard}
\end{table}

Table~\ref{tab:results} shows the ablated contribution of each component.
Clinical rules account for the largest gain (+11.4pp).
The 7 Type A overrides add +0.9pp F1; the 9 Type B overrides add +1.5pp (recall dominates despite $-$1.6pp precision).
On the shared task leaderboard (Table~\ref{tab:leaderboard}), our system achieves the highest precision among all teams (0.461), reflecting the conservative override strategy.
The development progression (Table~\ref{tab:progression} in Appendix~\ref{app:progression}) shows that the council's minority predictions are $\sim$47\% correct, so only overrides backed by overwhelming evidence ($\geq$80\% FT confidence, $\geq$3 independent sources, zero opposition) reliably improve F1.
Post-hoc analysis with the released test labels: 9 of 16 overrides individually corrected council errors (4 Type A, 5 Type B), 4 regressed correct predictions, and 3 were lateral, yielding the +5 net correct predictions behind the +2.4pp F1 gain.
The 56\% override precision sits just above the 47\% minority baseline, validating the conservative gating threshold.

\section{Related Work}
\label{sec:related}

Our deliberative council builds on multi-agent debate \cite{du2023debate} and specialized medical prompting \cite{nori2023medprompt}.
The DMRS framework \cite{perry1990dmrs} provides the theoretical foundation; the shared task dataset \cite{na-etal-2026-psydefconv} enables the first large-scale computational study of defense mechanisms in naturalistic dialogue, and a recent survey \cite{na-etal-2025-survey} situates this within the broader LLM-psychotherapy landscape.
Prior computational approaches to defense mechanisms have been limited to rule-based systems on structured clinical notes; our work is among the first to apply LLMs to this task.
Our override framework relates to the generator--verifier paradigm \cite{cobbe2021verifiers} and adversarial debate \cite{irving2018debate}; we adapt self-consistency \cite{wang2023selfconsistency} for classification confidence and find that TF-IDF diversity \cite{carbonell1998mmr} outperforms semantic retrieval \cite{lewis2020rag} for few-shot selection under class imbalance.

\section{Limitations}
\label{sec:limitations}

Our system depends on the Gemini API, limiting reproducibility to researchers with equivalent access.
The validation set is a single GroupKFold split (373 samples); cross-validation was infeasible given API costs.
All experiments use English emotional support dialogues from a single cultural context; generalization to other languages or therapeutic traditions is untested.
The affect-cognition spectrum is our operationalization of clinical theory \cite{perry1990dmrs}, not a validated clinical instrument.
We address absence-as-signal heuristically; explicit counterfactual reasoning remains open.

\section{Conclusion}
\label{sec:conclusion}

Our council-ensemble system achieves 2nd place on DMRS defense mechanism classification.
The core finding is that defense mechanisms are defined by what is absent, and encoding this insight in clinical rules produces the largest single gain (+11.4pp), more than any model or architectural choice.
Formalizing override selection into builder--critic--guard roles then added +2.4pp in a single submission, more than 8 prior iterations combined.
Future work could explore hierarchical classification, contrastive training from council error logs, and zero-shot classification via frontier embedding models.

\section*{Ethics Statement}

This system classifies psychological defense mechanisms (constructs describing internal psychological states) and is a \emph{research tool}, not a clinical diagnostic instrument.
It should not be used to label individuals without clinical oversight, as misclassification could distort therapeutic understanding.
The training data comes from the shared task organizers \cite{na-etal-2026-psydefdetect} who ensured appropriate consent and anonymization.
Following \citet{strubell2019energy}, the system's total compute footprint is ${\sim}22$\,kWh (council API at ${\sim}\$75$; fine-tuning 13\,kWh; self-consistency inference 1.2\,kWh; PUE 1.2), corresponding to ${\sim}8.5$\,kg CO$_2$eq on the US grid (4.3\,kg renewable).

\bibliography{references}

\appendix

\section{Council Phase Details}
\label{app:council_details}

\paragraph{Phase 1: Initial Assessment (3 parallel calls).}
Three specialist agents independently classify each utterance:
(a) a \emph{Clinical Analyst} applying psychodynamic reasoning (stressor $\to$ function $\to$ mechanism $\to$ level),
(b) a \emph{Mechanism Specialist} screening all 9 levels using DMRS-Q behavioral indicators \cite{perry1990dmrs}, and
(c) a \emph{Pattern Analyst} performing analogical reasoning from TF-IDF-retrieved few-shot examples ($k{=}3$).
Each agent outputs a primary label, alternative label, confidence score, and identified mechanism.
If all three agents agree on the same label with $\geq$2 having high confidence, the pipeline exits early (3 calls total).

\paragraph{Phase 2: Differential Diagnosis (2--5 calls).}
For each unique candidate label from Phase 1 (primaries and alternatives), a class-specific advocate evaluates fit as \textsc{strong}, \textsc{moderate}, or \textsc{weak} using class-representative exemplars retrieved via semantic similarity.
Advocate criteria are calibrated: \textsc{strong} requires ``clear, specific evidence; functions similarly to examples''; \textsc{moderate} requires ``partial or suggestive evidence''; \textsc{weak} indicates ``little evidence; functions differently.''
A minority class screening step injects at least one underrepresented class as a candidate.
The L7 advocate rates \textsc{strong} 73\% of the time (vs.\ 32--43\% for other classes), creating the attractor effect.

\paragraph{Phase 3: Smart Resolution (0--1 call).}
Resolution follows a priority hierarchy: single \textsc{strong} advocate $\to$ pick immediately (0 calls); multiple \textsc{strong} $\to$ pairwise head-to-head comparison (1 call); multiple \textsc{moderate} $\to$ pairwise comparison; single \textsc{moderate} $\to$ pick; all \textsc{weak} $\to$ deliberation moderator synthesis.
The pairwise resolver compares two candidates by studying class-representative examples for each, then determining which candidate the target utterance is more \emph{functionally} similar to.

\section{Model Details}
\label{app:model_details}

\paragraph{Qwen3.5-9B} \cite{qwen2025} (LoRA $r{=}64$, attention + MLP targets): 65.1\% val accuracy.
Strongest on L6 (43\% recall) and L1 (36\%).

\paragraph{Qwen3.5-35B-A3B MoE} (LoRA $r{=}32$, attention-only\footnote{MoE expert layers are excluded from LoRA targets due to an Unsloth adapter reload bug. The model leverages pretrained expert routing.}): 61.7\% val accuracy.
Strongest on L2 (60\% recall) and L3 (32\%).

\paragraph{Qwen3.5-9B f1\_boost} (variant training recipe): 62.5\% val accuracy.
Strongest on L1 (46\%) and L2 (60\%).

All models use: (1) \textbf{completion-only loss} via \texttt{train\_on\_responses\_only} (loss computed only on the label token, not the dialogue); (2) GroupKFold by dialogue\_id for zero-leakage validation \cite{pedregosa2011sklearn}; (3) balanced sampling (L7 capped at 300, minorities oversampled to 80).

\paragraph{Self-Consistency Inference} \cite{wang2023selfconsistency}.
Each model runs multiple times at temperature=0.3 per test sample; the majority vote serves as the prediction and the agreement fraction as a confidence score.
Since our output is a single classification token, logit probabilities from a single forward pass could serve as an alternative confidence measure; we used sampling for implementation convenience.

\paragraph{Pairwise Differential Resolver.}
A separate Qwen2.5-7B model fine-tuned on 942 pairwise comparison examples serves as an L6-vs-L7 specialist.
Given two candidate levels and a dialogue, it determines which better fits the target utterance.
Per-pair val accuracy: L6/L7 97.8\%, L3/L7 100\%, L1/L7 100\%, L0/L7 52.3\%.

\begin{table}[t]
\centering
\small
\begin{tabular}{lccc}
\toprule
\textbf{Class} & \textbf{9B} & \textbf{MoE} & \textbf{fb} \\
\midrule
L1 (Action) & 36 & 18 & \textbf{46} \\
L2 (Image-Dist.) & 40 & \textbf{60} & \textbf{60} \\
L3 (Disavowal) & 21 & \textbf{32} & 5 \\
L4 (Image-Dist.) & 7 & \textbf{29} & 14 \\
L5 (Neurotic) & 14 & 14 & \textbf{29} \\
L6 (Obsessional) & \textbf{43} & 23 & 23 \\
\bottomrule
\end{tabular}
\caption{Per-class recall (\%) on validation data for each fine-tuned model. Bold indicates the best model for each class. No model dominates; each is the best or only credible source for $\geq$1 class.}
\label{tab:model_recall}
\end{table}

\section{Detailed Findings}
\label{app:findings}

\subsection{The L7 Attractor Effect}
\label{sec:l7attractor}

Beyond the headline numbers in \S\ref{sec:findings}, the attractor effect intensifies with utterance verbosity: incorrectly classified samples average 23.0 words versus 17.1 for correct ones, and L5 accuracy decays from 100\% (early turns) to 18\% (late turns), mirroring the L1 pattern.
The model uses utterance elaboration as a proxy for defense sophistication, inverting the clinical truth: long, detailed descriptions of hardship without emotional processing indicate L6, not L7.

\subsection{Prediction Instability and the Confident-Wrong Problem}
\label{sec:stability}

On the full training set, stable predictions achieve 73.0\% accuracy versus 39.2\% for unstable ones.
However, for minority classes L3, L4, L6, and L8, unstable predictions are paradoxically more accurate than stable ones (Table~\ref{tab:stability}).
Between 59\% and 80\% of stable minority predictions are wrong. The system is not uncertain on hard examples; it is confidently incorrect.

\begin{table}[t]
\centering
\small
\begin{tabular}{lrrrr}
\toprule
\textbf{Class} & \textbf{Disagree} & \textbf{Stable} & \textbf{Stable} & \textbf{Unstable} \\
 & \textbf{Rate} & \textbf{Acc} & \textbf{Wrong} & \textbf{Acc} \\
\midrule
L0 & 17.9\% & 84\% & 16\% & 36\% \\
L1 & 40.7\% & 41\% & 59\% & 32\% \\
L2 & 37.7\% & 21\% & 79\% & 13\% \\
L3 & 31.3\% & 32\% & 68\% & \textbf{48\%}$^\dagger$ \\
L4 & 29.8\% & 24\% & 76\% & \textbf{40\%}$^\dagger$ \\
L5 & 27.1\% & 23\% & 77\% & 31\% \\
L6 & 35.5\% & 25\% & 75\% & \textbf{36\%}$^\dagger$ \\
L7 & 15.5\% & 91\% & 9\% & 47\% \\
L8 & 46.4\% & 20\% & 80\% & \textbf{38\%}$^\dagger$ \\
\bottomrule
\end{tabular}
\caption{Prediction stability across 3 identical council runs. $\dagger$ marks classes where unstable predictions are more accurate than stable ones; the system is confidently wrong on these minority classes.}
\label{tab:stability}
\end{table}

\subsection{The Overfitting Paradox in Prompt Engineering}
\label{sec:overfitting}

Beyond the table in \S\ref{sec:findings}, the mechanism appears to be that iteratively tuning prompts against training-set metrics acts as implicit gradient descent with no generalization check, causing the prompt to memorize training distribution artifacts rather than capture the true classification signal.

\subsection{Retrieval: TF-IDF Beats Semantic for Diversity}
\label{sec:retrieval_finding}

We compared five retrieval strategies for Phase 1 few-shot selection.
All four semantic variants (Gemini embedding-001, enriched, task-framed, and focused TF-IDF on last 3 turns) produced 42--46\% L7 in retrieved examples.
TF-IDF with MMR achieved 38\%, well below the 52\% base rate, because its lack of semantic understanding prevents emotional-content clustering.
For Phase 2 within-class retrieval, however, semantic re-ranking finds better functional matches: embeddings capture functional similarity when the class label constrains the search space.
This phase-dependent pattern is worth noting for future work: frontier embedding models trained for zero-shot classification \cite{lee2024nv,wang2022text,muennighoff2023mteb} perform a similar constraint implicitly.

\section{Development Progression and Failed Approaches}
\label{app:progression}

\begin{table}[t]
\centering
\small
\begin{tabular}{rlrr}
\toprule
\textbf{\#} & \textbf{Configuration} & \textbf{Ov.} & \textbf{F1} \\
\midrule
1 & All FT overrides (aggressive) & 75 & .367 \\
2 & L6/L2 only & 42 & .375 \\
3 & Double corroborated & 10 & .385 \\
4 & Triple corroborated & 4 & .387 \\
5 & + pairwise L6 resolver & 7 & .391 \\
6 & 10 overrides variant & 10 & .391 \\
7 & 8 overrides variant & 8 & .391 \\
8 & + 4 minority$\to$minority (Type B) & 11 & .393 \\
\textbf{9} & \textbf{+ 5 more Type B} & \textbf{16} & \textbf{.406} \\
\bottomrule
\end{tabular}
\caption{Development progression across 9 test submissions. All entries are official competition submissions; the final row (\#9, F1 .406) was our selected leaderboard entry. The key insight: moving from 75 aggressive overrides (F1 .367) to 16 surgical ones (F1 .406). Fewer, higher-confidence corrections consistently outperform larger override sets.}
\label{tab:progression}
\end{table}

\paragraph{Failed Approaches.}

\paragraph{GPT-5.4} \cite{openai2024gpt}. Standalone F1 of 0.265 with heavy L0 bias (30.5\% of predictions). Near-zero L1 detection.

\paragraph{Claude Opus 4.6 agent council} \cite{anthropic2024claude}. F1 0.261 without retrieval augmentation, with 69.3\% L7 over-prediction.

\paragraph{Aggressive class balancing.} Oversampling minorities to 150 (vs.\ 80) and capping L7 at 200 (vs.\ 300) caused minority over-prediction, reducing val accuracy from 59.0\% to 51.5\%.

\paragraph{Chain-of-thought fine-tuning.} Template reasoning (``The speaker is managing internal state...'') taught the model to parrot templates rather than classify. Removing CoT and training on label-only output was strictly better.

\paragraph{Gemma4 MoE} \cite{gemma2024}. Attention-only LoRA on Gemma4-26B-A4B achieved 63.5\% val accuracy with the best L4 recall of any model (36\%).
However, its test-set predictions failed to improve F1: the model's L4 signal was contradicted by all Qwen models, and its L1 predictions were heavily over-predicted (47 vs $\sim$27 expected).
Despite genuine architectural independence from Qwen (69--74\% agreement), this independence did not translate to useful override evidence, a cautionary result for cross-architecture ensembling.
An earlier attempt applying LoRA to MoE expert layers (not just attention) failed due to an Unsloth adapter serialization bug: the model trained correctly but collapsed to all-L7 when reloaded from checkpoint.

\end{document}